\documentclass[10pt,twocolumn,letterpaper]{article}

\usepackage[pagenumbers]{cvpr} 

\usepackage{multirow}
\usepackage{scalerel}  

\usepackage{multirow}
\usepackage{colortbl}
\usepackage[normalem]{ulem}
\useunder{\uline}{\ul}{}

\usepackage[dvipsnames]{xcolor}

\definecolor{cvprblue}{rgb}{0.21,0.49,0.74}
\usepackage[pagebackref,breaklinks,colorlinks,allcolors=cvprblue]{hyperref}
\usepackage[inkscapelatex=false]{svg}
\def\confName{CVPR}
\def\confYear{2025}
\usepackage{marvosym}

\def\logo{\scalerel*{\includegraphics{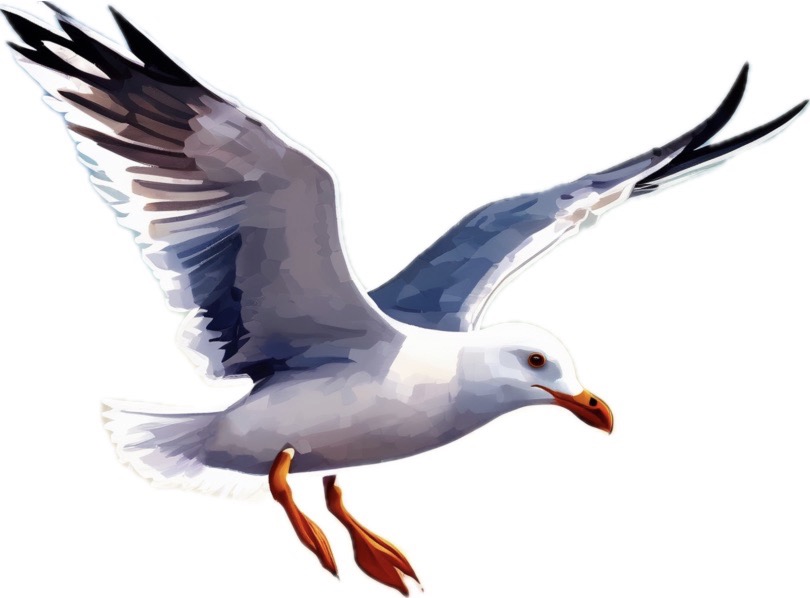}}{\textrm{\textbigcircle}}}
\newcommand{\titledmodelname}{\textsc{Seagull}\logo\xspace}
\title{\titledmodelname: No-reference Image Quality Assessment for Regions of Interest via Vision-Language Instruction Tuning}

\author{Zewen Chen$^{1,2}$, Juan Wang$^{1}$, Wen Wang$^{3}$, Sunhan Xu$^{4}$, Hang Xiong$^{3}$, Yun Zeng$^{5}$, Jian Guo$^{4}$, \\Shuxun Wang$^{1}$, Chunfeng Yuan$^{1}$, Bing Li$^{1,6}$\textsuperscript{\Letter}, Weiming Hu$^{1,2,7}$ \\
$^1$ State Key Laboratory of Multimodal Artificial Intelligence Systems, CASIA \\
$^2$ School of Artificial Intelligence, University of Chinese Academy of Sciences \\
$^3$ Beijing Jiaotong University $^4$ Beijing Union University $^5$ China University of Petroleum \\
$^6$ PeopleAI Inc. Beijing, China \\
$^7$ School of Information Science and Technology, ShanghaiTech University \\
\{chenzewen2022, jun\_wang\}@ia.ac.cn, bli@nlpr.ia.ac.cn \\
}

\begin{document}
\maketitle
\begin{abstract}
Existing Image Quality Assessment (IQA) methods achieve remarkable success in analyzing quality for overall image, but few works explore quality analysis for Regions of Interest (ROIs). The quality analysis of ROIs can provide fine-grained guidance for image quality improvement and is crucial for scenarios focusing on region-level quality. This paper proposes a novel network, \textsc{Seagull}, which can \textbf{SE}e and \textbf{A}ssess ROIs quality with \textbf{GU}idance from a \textbf{L}arge vision-\textbf{L}anguage model. \textsc{Seagull} incorporates a vision-language model (VLM), masks generated by Segment Anything Model (SAM) to specify ROIs, and a meticulously designed Mask-based Feature Extractor (MFE) to extract global and local tokens for specified ROIs, enabling accurate fine-grained IQA for ROIs. Moreover, this paper constructs two ROI-based IQA datasets, \textsc{Seagull}-100w and \textsc{Seagull}-3k, for training and evaluating ROI-based IQA. \textsc{Seagull}-100w comprises about 100w synthetic distortion images with 33 million ROIs for pre-training to improve the model's ability of regional quality perception, and \textsc{Seagull}-3k contains about 3k authentic distortion ROIs to enhance the model’s ability to perceive real world distortions. After pre-training on \textsc{Seagull}-100w and fine-tuning on \textsc{Seagull}-3k, \textsc{Seagull} shows remarkable performance on fine-grained ROI quality assessment. Code and datasets are publicly available at the \href{https://github.com/chencn2020/Seagull}{link}.
\end{abstract}    
\section{Introduction}
\label{sec:intro}

Image quality assessment (IQA) is a long-standing research in image processing fields. 
Compared to the full-reference IQA (FR-IQA) and the reduced-reference IQA, the no-reference IQA (NR-IQA) receives more attention since the reference images are unavailable in real-word applications. 

\begin{figure}[htbp]
    \centering
    \includegraphics[width=\columnwidth]{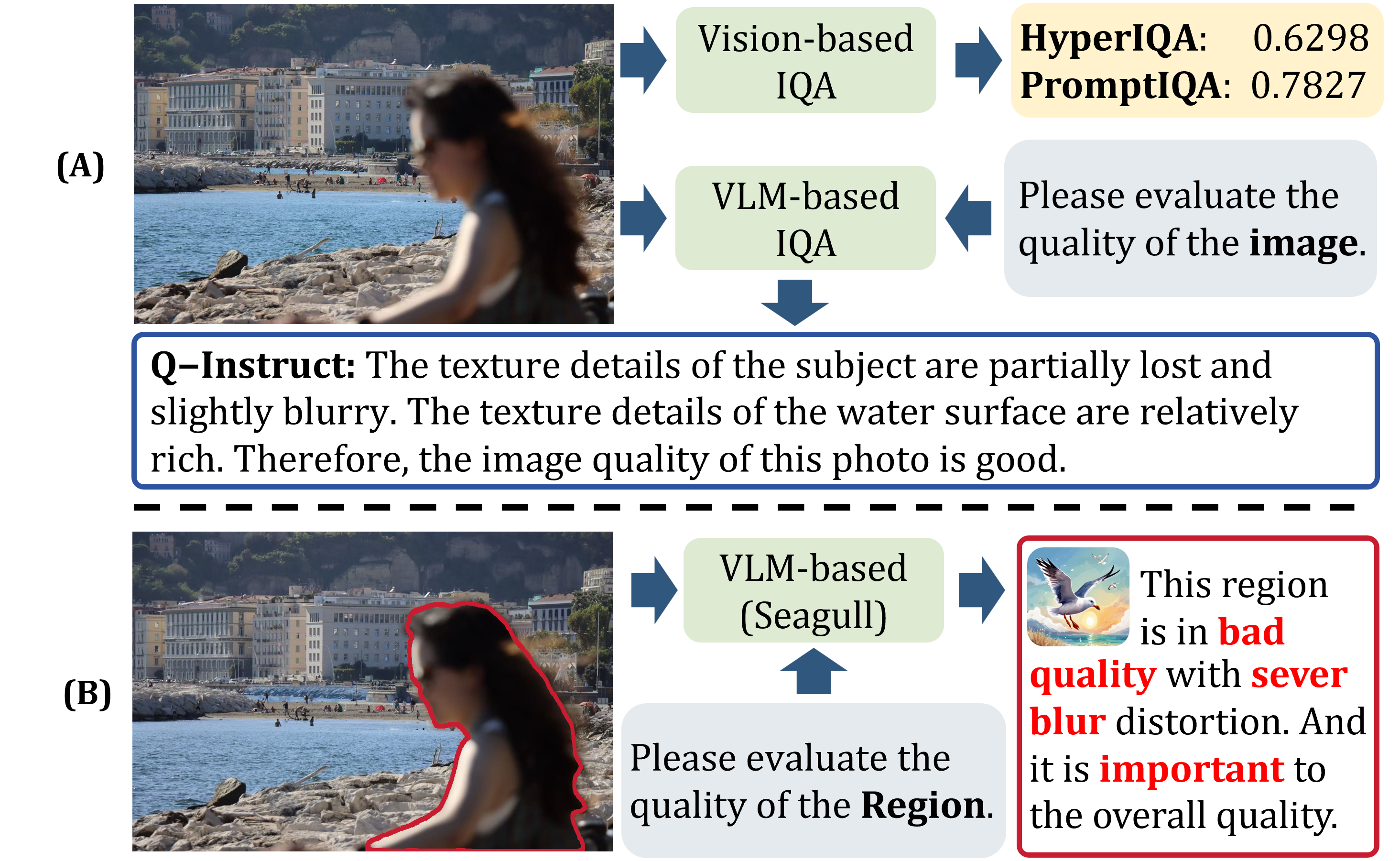}
    \caption{(A) Illustrations of the typical Vision-based and VLM-based IQA. Both of them are designed to analyze the quality of overall image. (B) Our \textsc{Seagull} has the capability in fine-grained quality assessment for specified ROI. The mask-based ROI is extracted by SAM \cite{SAM}. Best viewed in color. 
    }
    \label{fig:idea}
\end{figure}

As illustrated in Fig.\ref{fig:idea}-(A), NR-IQA methods can be categorized into two main types based on their inputs: Vision-based IQA \cite{ke2021musiq, yang2022maniqa, chen2024promptiqa, wang2023Hierarchical, su2020blindly} and Vision-Language Model (VLM)-based IQA \cite{wu2023qinstruct, wu2024qbench, you2024descriptive, you2023depicting}. Vision-based IQA predicts quality scores for the input images. This approach lacks of interpretability. VLM-based IQA provides detailed quality descriptions based on the input images and textual prompts.
This approach has achieved a significant success in analyzing quality for overall images, but overlooks the exploration of quality assessment for specified Regions of Interest (ROIs).
Analyzing the quality for ROIs provides more refined guidance for image quality improvement, which has wide applications across various domains, such as video optimization for focused objects \cite{Kong2016Video, Kwon2010videoROI,Ling2009ROI,kim2021ROIAssess}, image enhancement for ROIs \cite{Kaur2016CICT,yang2018roi,gao2023reversible,kim2021assessment,wang2023micro} and image compression for interested regions  \cite{ma2021variable, 4107225, kao2023transformer}.  

One method to achieve the quality assessment for ROIs is cropping ROIs and then directly feeding cropped regions into existing vision-based IQA models. Alternatively, drawing bounding boxes (BBoxes) on images can also be used to indicate ROIs.
However, both crop-based ROIs and BBox-based ROIs include irrelevant background and struggle to precisely indicate interested objects, leading to inaccurate instructions to IQA models. In contrast, using masks to indicate ROIs (mask-based ROIs) accurately delineates the focused regions.
Recently, numerous segmentation models \cite{SAM, NEURIPS2023_3ef61f7e, ke2024segment} exhibit excellent zero-shot performance and many advanced VLMs \cite{achiam2023gpt,liu2024visual,ye2023mplug} show impressive capabilities in visual understanding.  These advancements make it potential to achieve IQA for ROIs by utilizing segmentation models to generate masks that indicate ROIs and VLMs for fine-grained ROI quality assessment.
Nevertheless, there are two challenges: (1) The performance of these VLMs in quality analysis is limited \cite{wu2024qbench, wu2023qinstruct, you2024descriptive}, since they are designed and trained for high-level tasks and ignore to effectively extract low-level features. (2) Most existing IQA datasets \cite{sheikh2006statistical,ciancio2010no,hosu2020koniq,wu2023qinstruct} only provide quality scores or descriptions for the overall images, which make them unsuitable for  ROI-based IQA training. 

In this paper, we achieve fine-grained IQA for ROIs from the perspective of network and dataset. Firstly, we design a novel network that consists of the Segment Anything Model (SAM) \cite{SAM} to extract mask-based ROIs for accurately \textbf{seeing} and a VLM to comprehend the ROIs' quality for fine-grained \textbf{assessing}. We name this network \textbf{\textsc{Seagull}}, which can \textbf{\textit{SE}}e and \textbf{\textit{A}}ssess ROIs quality with \textbf{\textit{GU}}idance from a \textbf{\textit{L}}arge vision-\textbf{\textit{L}}anguage model. Additionally, to enhance the capacity of \textsc{Seagull} in ROIs assessing, we meticulously design a mask-based feature extractor (MFE). MFE extracts global and local view tokens using mask-based ROIs, providing \textsc{Seagull} with more perspectives to understand the quality of the ROIs. At last, by clicking any region on a full image, SAM first extracts the mask to indicate the ROI, and \textsc{Seagull} subsequently performs a fine-grained quality assessment on the ROI using the extracted mask.

Secondly, we construct two ROI-based IQA datasets, namely \textbf{\textsc{Seagull}-100w} and \textbf{\textsc{Seagull-3k}}. Both of them provide labels for ROIs from three dimensions: ROI Quality Score, ROI Importance Score, and ROI Distortion Analysis. \textbf{ROI Quality Scores} quantitatively measure the quality for a specific ROI. \textbf{Importance Scores} reflect the impact of an ROI on the overall image. \textbf{Distortion Analysis} provides distortion types and the severity degree. Specifically, \textsc{Seagull}-100w, consisting of about 100w synthetic distortion (Dist.) images and approximately 33 million mask-based ROIs, is constructed for pre-training to enhance the quality perception ability of models. Each ROI is carefully annotated with the three dimensions labels using reliable models. Specially, Dist. images in \textsc{Seagull}-100w are derived from the RAW images using an Image Signal Processor (ISP) under different distortions settings. Compared to RGB-derived Dist. images, RAW-derived Dist. images are more authentic, since more detailed information captured from the camera sensor is reserved in the RAW. 

However, the discrepancy between synthetic and authentic Dist. image hinders the robustness of models in real-world images. Thus, we additionally construct the manually annotated \textbf{\textsc{Seagull}-3k} dataset for fine-tuning. This dataset consists of 3,261 ROIs for real-world authentic Dist. images and is annotated by 24 trained annotators. For each ROI, annotators are required to evaluate the distortion types, distortion severity degrees, quality score of the ROI and its importance on the overall image quality. To reduce bias, each ROI is annotated by at least 7 annotators, and the labels are determined based on their average results.

As shown in Fig.\ref{fig:idea}-(B), after pre-training on \textsc{Seagull}-100w and fine-tuning on \textsc{Seagull}-3k,
\textsc{Seagull} achieves fine-grained quality analysis for the specified ROI. 
Extensive experiments demonstrate the superiority of \textsc{Seagull}. Our contributions can be summarized as follows:
\begin{enumerate}
    \item We propose \textsc{Seagull}, a novel IQA framework that accomplishes fine-grained quality assessment for any specified ROIs from the aspects of Quality Score, Importance Score and Distortion Analysis.
    \item We construct two ROI-based IQA datasets: \textsc{Seagull}-100w and \textsc{Seagull}-3k. Compared to existing datasets, ours provide more detailed labels for ROI-based IQA.
    \item Experimental results demonstrate that \textsc{Seagull} remarkably surpasses existing advanced IQA models and VLMs in ROI quality analysis.

\end{enumerate}
\section{Related Work}
\label{sec:related_work}

\begin{figure*}[htbp]
  \includegraphics[width=\textwidth]{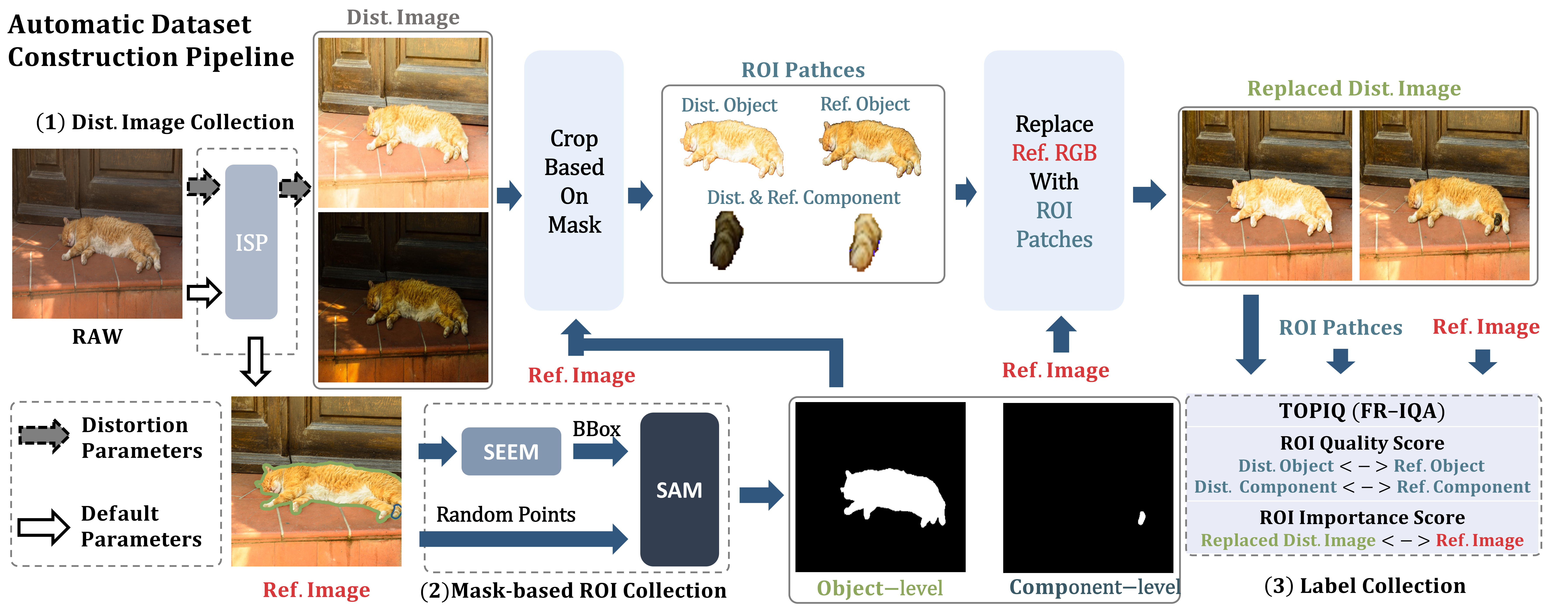}
  \caption{The automatic pipeline for generating the \textsc{Seagull}-100w dataset.}
  \label{fig:pipeline}
\end{figure*}

\subsection{Vision-based IQA}

Vision-based IQA models solely rely on vision features extracted from input images to predict quality scores. Recently, numerous studies improve Vision-based IQA from various perspectives, such as proposing efficient network architectures \cite{kang2014convolutional,ke2021musiq,yang2022maniqa,xu2023local,qin2023data}, introducing relevant proxy tasks for pre-training \cite{chen2022teacher,lin2018hallucinated,wang2021active,pan2022vcrnet}, proposing mixed training methods \cite{wang2023Hierarchical,chen2024promptiqa,sun2023blind,zhang2018blind} to extract universal information from diverse IQA datasets.
Additionally, techniques such as negative sample mining \cite{wang2023deep}, efficient loss function \cite{ li2020norm}, quality ranking learning \cite{golestaneh2022no,gao2015learning,liu2017rankiqa}, and contrastive learning \cite{zhao2023quality,madhusudana2022image,shi2023transformerbased} have also been explored to further enhance the performance and robustness of Vision-based IQA models.
Although these models show powerful performance, they only predict a single overall quality score, which poses challenges to the interpretability of IQA.

\subsection{Vision-Language-based IQA}
Recently, many VLMs \cite{achiam2023gpt,alayrac2022flamingo,NEURIPS2023_6dcf277e,ye2023mplug,yuan2024osprey,dai2024instructblip} achieve powerful performance on visual understanding and also show potential in quality perception \cite{wu2024qbench, wu2023qinstruct, you2024descriptive}. 
For example, Wu \etal propose Q-Bench \cite{wu2024qbench} and Q-Instruct\cite{wu2023qinstruct} to evaluate and enhance the quality reasoning capabilities of VLMs. Zhang \etal \cite{zhang2023blind} propose an all-in-one IQA model using a language encoder. Moreover, Q-Align \cite{wu2023qalign} integrates Image Aesthetic Assessment, IQA and Video Quality Assessment in one VLM. You \etal \cite{you2024descriptive,you2023depicting} explore the ability of VLMs in IQA reasoning and pair-wise comparison. Although VLM-based IQA models demonstrate potential in overall quality analysis and scores predicting, they all overlook the exploration of assessing  quality for ROIs.

\subsection{IQA Datasets}

Most existing IQA datasets \cite{sheikh2006statistical,larson2010most,ponomarenko2013color,kadid10k,ciancio2010no,fang2020perceptual,ghadiyaram2015massive,hosu2020koniq} only provide overall quality scores as labels, lacking fine-grained annotation for ROIs. FLIVE \cite{Ying_2020_CVPR} provides quality scores for patches. However, these patches are randomly cropped and therefore do not contain practical semantics. Moreover, some reasoning IQA datasets \cite{wu2024openended, wu2023qinstruct,you2024descriptive,wu2024qbench} have been proposed to improve the interpretability of IQA models. However, they only provide the coarse description on the overall images and lack the fine-grained description on the specified ROIs. Chen \etal construct QGround-100K \cite{chen2024q} with mask-based ROIs. The dataset provides a single distortion label for each ROI, while ROIs in real-world images often exhibit mixed distortions or high quality. Additionally, it does not include other fine-grained labels about distortion severity degree and the importance of ROIs, which are essential for fine-grained assessment.

Although existing methods improve IQA performance and interpretability from various aspects, they mainly focus on overall IQA and neglect the exploration of ROI-based IQA. To address the challenges of unsuitable networks and datasets, we propose a novel network (\textsc{Seagull}) and construct two ROI-based IQA datasets (\textsc{Seagull}-100w and \textsc{Seagull}-3k) to achieve IQA for ROIs.

\section{The Proposed Datasets}

In this section, we introduce the construction process of the \textsc{Seagull}-100w and \textsc{Seagull}-3k. To make the two datasets suitable for VLMs tuning, we introduce the details of instruction-response design.

\subsection{\textbf{\textsc{Seagull}}-100w Dataset}
\label{sec:dataset}

Fig.\ref{fig:pipeline} shows three stages in the automatic  construction pipeline for \textsc{Seagull}-100w: (1) \textbf{Dist. image Collection} for large-scale synthetic data generating, (2) \textbf{Mask-based ROI Collection} for rich and diverse ROIs obtaining, and (3) \textbf{Label collection} for ROIs fine-grained quality annotations generating. Next, we introduce the three stages in detail.

\subsubsection{Distortion Image Collection}
\label{sec:dist_img_collection}

Existing synthetic IQA datasets \cite{sheikh2006statistical,larson2010most,ponomarenko2013color,kadid10k} generate Dist. images using RGB images. However, RGB images significantly compress the sensor captured information, leading to the reduction of details. This makes RGB-derived Dist. images less authentic. For this reason, we take a professional ISP, Adobe LightRoom (LR), to generate Dist. images using 8,146 RAW images in RAISE \cite{10.1145/2713168.2713194}, since RAW images retain all sensor data, offering more details and higher dynamic ranges. 

Specifically, we firstly generate reference (Ref.) images with the default parameters of cameras. Then, by adjusting the parameters of different ISP modules, we produce six types of distortions, namely \textit{\textbf{exposure}}, \textit{\textbf{noise}}, \textit{\textbf{blur}}, \textit{\textbf{contrast}}, \textit{\textbf{colorfulness}}, and \textit{\textbf{compression}}. 
We sample twenty different parameters for each of the six distortion types to  generate Dist. images. Finally, we obtain about 100w Dist. images.

\subsubsection{Mask-based ROI Collection}
\label{sec:mask_collection}
With the advancements in open-world segmentation models such as SAM \cite{SAM} and SEEM \cite{NEURIPS2023_3ef61f7e}, it is efficient and reliable to obtain mask-based ROIs. 
As shown in Fig.\ref{fig:pipeline}, we define \textbf{object-level} and \textbf{component-level mask} to assess the quality for an entire object (e.g., a cat) or a component of an object (e.g., the tail of a cat), respectively.

For object-level masks, we firstly employ SEEM \cite{NEURIPS2023_3ef61f7e} to detect objects in Ref. images. Subsequently, we employ SAM to extract object-level masks based on these detection BBoxes. For component-level masks, we directly adopt SAM to capture component-level masks with random points for each Ref. image and filter out small ROIs whose areas are smaller than $32\times32$ pixels. At last, we obtain about 10 million object-level and 23 million component-level masks.

\subsubsection{Labels Collection}
\label{sec:label_collection}
We annotate three types of labels to indicate the ROI's quality, namely 
\textbf{ROI Quality Scores}, \textbf{ROI Importance Scores} and \textbf{ROI Distortion Labels}.

\textbf{A) ROI Quality Score.} 
Firstly, we crop Dist. and Ref. images into patches using mask-based ROIs collected in Sec.\ref{sec:mask_collection} and pad them into minimum rectangles.
Then, we obtain quality scores for Dist. ROIs using TOPIQ \cite{10478301}, a state-of-the-art (SOTA) FR-IQA that improves the performance by combining semantic and distortion features in a top-down approach.

\textbf{B) ROI Importance Score.}
Different ROIs have significantly different impacts on the overall quality.
Improving the quality of critical ROIs effectively enhances the overall quality of an image \cite{2012Enhanced}.
Thus, we introduce importance scores to assess the importance of ROIs to the overall image quality.
Specifically, we directly replace the corresponding position of the Ref. image with the Dist. ROI to obtain the replaced Dist. image, and then evaluating the difference between the replaced Dist. image and the Ref. image using TOPIQ.
If the ROI is important to the overall quality, there will be a significant score difference between the two images. Hence, we adopt this method to measure the importance of ROIs to the overall image quality.

\textbf{C) ROI Distortion Labels.}
As described in Sec.\ref{sec:dist_img_collection}, we create the Dist. images with different distortion types. The applied single distortion is set as the distortion label for each ROI.
Moreover, we find that ROIs with quality scores higher than 0.92 show no much difference from the Ref. images. Thus, we set the distortion labels of these ROIs with the scores above 0.92 to ``Without distortions''. Otherwise, it is labeled as ``with'' the corresponding distortion. 

\subsection{\textbf{\textsc{Seagull}}-3k Dataset}
Although synthetic Dist. images with reliable annotations are easily accessible, they differ significantly from authentic images, which often display multiple and uneven distortions. This discrepancy hinders the robustness of models in real-world images. 
Thus, we additionally construct \textsc{Seagull}-3k dataset, comprising authentic Dist. images with ROI labels annotated by 24 trained annotators. 

Initially, we collect 968 Dist. images from four existing authentic IQA datasets: LIVEC \cite{ghadiyaram2015massive}, BID \cite{ciancio2010no}, SPAQ \cite{fang2020perceptual}, and KonIQ \cite{hosu2020koniq}. Subsequently, following Sec.\ref{sec:mask_collection}, we obtain two or three component-level and object-level ROIs for each images and obtain 3,261 ROIs. 
For each ROI, at least 7 annotators assign the labels. Specifically, for the distortion label, annotators assess the six specified distortions and assign a rating from six pre-defined distortion levels: extreme (0), severe (1), moderate (2), minor (3), trivial (4), or non-existent (5). If a majority of annotators rate ``non-existent", then the ROI is labeled as ``without" that specific distortion; conversely, it is labeled as ``with" the distortion and its severity degree is calculated by averaging the levels rated by annotators.  
For the quality label, annotators assign ROI quality from five levels: bad (0), poor (1), fair (2), good (3) and excellent (4). Similarly, ROI importance score is rated from unimportant (0), minor (1), normal (2), important (3) and essential (4). The final quality and importance score of a ROI are the average of scores assigned by annotators.

Subsequently, we discretize these scores into five categorical levels using the following equation \cite{wu2023qalign}:
\setlength{\abovedisplayskip}{3.5pt}
\setlength{\belowdisplayskip}{3.5pt}
\begin{equation}
K(s^t)=k^t_i \quad \text{if} \quad \frac{\mathrm{M}\cdot i}{5} < s^t \leq \frac{\mathrm{M}\cdot (i + 1)}{5},
\end{equation}
where $s^t$ denotes scores for quality ($s^{qs}$), importance ($s^{is}$) and distortion severity degree ($s^{dsd}$). $k^t_i \in \{k^{qs}_i, k^{is}_i, k^{dsd}_i\}$ represents categories [Bad, Poor, Fair, Good, Excellent] for quality scores ($k^{qs}_i$), [Unimportant, Minor, Normal, Important, Essential] for importance scores ($k^{is}_i$), and [Extreme, Severe, Moderate, Minor, Trivial] for distortion severity levels ($k^{dsd}_i$) and $i$ is the category index.

\begin{figure*}[htbp]
  \centering
  \includegraphics[width=\textwidth]{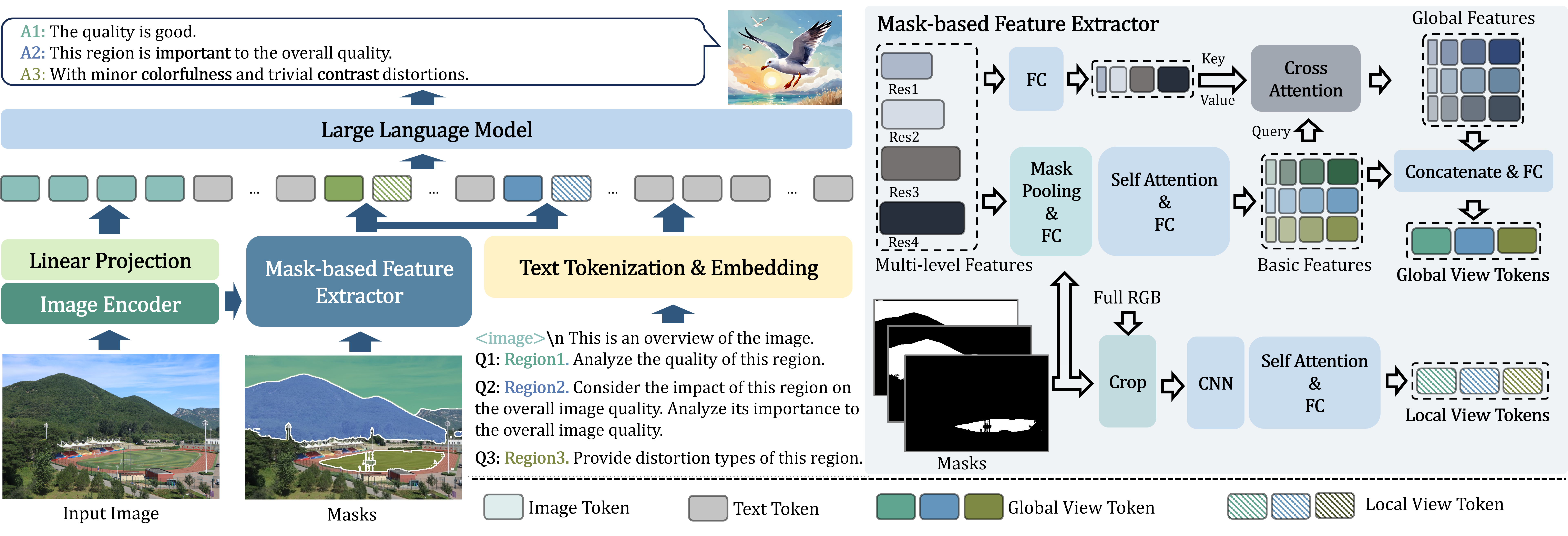}
  \caption{Overview of the \textsc{Seagull} (left) and the Mask-based Feature Extractor (right). Best viewed in color.}
  \label{fig:framework}
\end{figure*}

\subsection{Instruction-Response Design}
\label{sec:instruction_response_design}

It is crucial to design reliable instruction-responses for VLM tuning. 
In this paper, we design two types of instruction-response namely Analysis Instruction-Response (AIR) and Judgment Instruction-Response (JIR). The AIR encourages VLMs to answer the questions about the ROI quality, importance and distortion analysis. The JIR queries the model whether the provided instructions are correct.

\noindent\textbf{A) Analysis Instruction-Response.} 
We design three types of AIRs: 1) For ROI quality scores: ``Analyze the quality of this region''. 2) For ROI importance scores: ``Consider the impact of this region on the overall image quality. Analyze it's importance''.  3) For distortion labels: ``Analyze the distortions of this region''.

\noindent\textbf{B) Judgment Instruction-Response.} 
To enhance the quality understanding and robustness of  VLMs, we additionally design three JIRs, namely\textbf{ quality score judgment}, \textbf{importance score judgment} and \textbf{distortion type judgment}, with queries like: ``Is the quality of this region good'', ``Is this region unimportant to the overall quality" or ``Is this region in blur distortion''  to judge whether a ROI meets the asked conditions. The expected responses are ``Yes" or ``No".

\section{The Proposed \textbf{\textsc{Seagull}} Network}
\label{src:model}

As illustrated in Fig.\ref{fig:framework}, different from most VLMs \cite{wu2023qalign,liu2024visual} that primarily consist of an image encoder and a LLM, \textsc{Seagull} additionally incorporates a mask-based feature extractor (MFE) to extract fine-grained features for ROIs. In the following, we introduce these parts in detail.

\subsection{Image Encoder For Multi-level Perception}

It is essential to extract robust image tokens. 
CNN-based models such as ResNet \cite{he2016deep} and ConvNeXt \cite{liu2022convnet} not only accelerate the training and inference processes, but also extract both global and local features. These features are useful to IQA tasks \cite{chen2022teacher,su2020blindly}. In our work, we employe the ConvNeXt-Large model as the image encoder.
Specifically, for an image $I \in \mathbb{R}^{3 \times H \times W}$, the ConvNeXt-Large model sequentially processes it through four layers to extract multi-level image tokens, denoted as $\mathcal{F}_j \in \mathbb{R}^{C_j \times H_j \times W_j}$, where $C_j$, $H_j$ and $W_j$ represent the respective dimensions of the feature for $1\leq j \leq 4$. In \textsc{Seagull}, the image tokens from the final layer ($\mathcal{F}_4$) are utilized as image tokens and fed into a linear project to align with text tokens.

\subsection{Mask-based Feature Extractor}
\label{sec:RQSFE}
The mask-based feature extractor (MFE) extracts both global and local view tokens using mask-based ROIs. Global view tokens contains ROI-related quality and semantic features extracted from long distance contexts. And local view tokens provide a detailed and direct representation of these ROIs. Both of them supply \textsc{Seagull} with different perspectives to understand the quality of ROIs.

Specifically, for a binary mask $M \in \mathbb{R}^{1 \times H \times W}$, we firstly take the mask-pooling  $\mathbf{MP}(\cdot)$ \cite{xu2023open} to extract regional tokens from the multi-level image tokens that fall inside the mask region. Subsequently, these regional tokens are projected into a uniform dimension using full connection (FC) layers. We denote them as multi-level basic tokens, which can be formulated as follows:
\begin{equation}
    \mathcal{B}_{j} = \mathbf{FC}_j(\mathbf{MP}(M, \mathcal{F}_j)) \in \mathbb{R}^{1\times P},
\end{equation}
where $\mathbf{FC}_{j}(\cdot)$ represents the $j$-th FC layer and $P$ denotes the dimension and $1\leq j \leq 4$.

However, it is challenging for \textsc{Seagull} to achieve accurate ROI quality analysis solely based on basic tokens, since these tokens lack global contexts.
To this end, we use the cross-attention (CA) mechanism to assist \textsc{Seagull} to gain comprehensive understandings of ROIs by capturing long distance contexts. At first, we take a self-attention (SA) mechanism on the multi-level basic tokens $\mathcal{B}_{j}$ to make \textsc{Seagull} comprehend basic information. For multi-level image tokens, we adopt the reshape operation to change the original dimensions from $\mathbb{R}^{C_j\times H_j\times W_j}$ to  $\mathbb{R}^{C_j\times (H_j\times W_j)}$. Then, another FC layers project these features into the same dimension as the multi-level basic tokens, \ie $\mathcal{F}_j = \mathbf{FC}_j(\mathcal{F}_j) \in \mathbb{R} ^ {C_j \times P}$. Finally, we set the multi-level basic tokens as queries ($Q=\mathbf{SA}(\mathcal{B}_{j})$) to capture the global attention tokens $\mathcal{G}$ from these multi-level image tokens served as keys ($K=\mathcal{F}_j$) and values ($V=\mathcal{F}_j$)
, which can be formulated as follows:
\begin{equation}
\resizebox{.9\linewidth}{!}{$
    \mathcal{G}_{j} = \mathbf{CA}(Q=\mathbf{SA}(\mathcal{B}_{j}), K=\mathcal{F}_j, V=\mathcal{F}_j) \in \mathbb{R}^{1\times P},
$}
\end{equation}
where $\mathbf{CA}(\cdot)$ and $\mathbf{SA}(\cdot)$ represent the cross-attention and self-attention operation.
After summing up the 4 basic tokens $\mathcal{B}_{j}$ and 4 global tokens $\mathcal{G}_{j}$, respectively, two additional FC layers are used to further fuse the summed tokens. Then, we concatenate the two fused tokens into one combined feature, and feed it into another FC layer to gain the final global view tokens. This process can be formulated as follows:
\begin{equation}
\resizebox{.89\linewidth}{!}{$
    \begin{split}
        \mathcal{F}_{gvt} = \mathbf{FC}_1\left\{\mathbf{CAT}\left[\mathbf{FC}_2\left(\sum\nolimits_{i=1}^4\mathcal{B}_{j}\right), \mathbf{FC}_3\left(\sum\nolimits_{i=1}^4\mathcal{G}_{j}\right)\right]\right\},
    \end{split}
$}
\end{equation}
where $FC_{k=1,2,3}(\cdot)$ represents FC layers and $\mathbf{CAT}(\cdot)$ denotes the concatenation operation.

$\mathcal{F}_{gvt}$ represents the global information related to the ROIs. To make \textsc{Seagull} better comprehend the quality of the ROIs, it is also necessary to provide a local view. Although basic tokens contain local features, they offer limited details for ROIs. That is because basic tokens are extracted from multi-level image features, which represent the full image rather than the details of specified ROIs. Thus, we additionally introduce a local view token $\mathcal{F}_{lvt}$.
Specifically, we employ two CNNs to capture the features from the cropped ROI ($\mathcal{I}_c$) using masks. Then, we use a SA to further fuse the features and adopt a FC layer to project it into the same dimension as the $\mathcal{F}_{gvt}$. Finally, we  get the local view token $\mathcal{F}_{lvt}=\mathbf{FC}(\mathbf{SA}(\mathbf{CNN}(\mathcal{I}_c))) \in  \mathbb{R}^{1\times P}$.

\subsection{LLM For Multi-modal Features Understanding}
The LLM fuses different multi-modal features and complete tasks based on text sequences. We set the text sequences similar to ``	\textless{}image\textgreater{}\textbackslash{}n. This is the overview of the image. 
Here is the region \textless{}global\textgreater{} \textless{}local\textgreater{}. Analyze the quality of the region''. This sequence is tokenized into text tokens, where the three special placeholders (\textless{}image\textgreater{}, \textless{}global\textgreater{} and \textless{}local\textgreater{}) are replaced with image tokens ($\mathcal{F}_4$), global view tokens ($\mathcal{F}_{gvt}$) and local view tokens ($\mathcal{F}_{lvt}$).

\section{Experiment}
\label{sec:exp}

\subsection{Experiment Settings}
\noindent \textbf{A) Training Stages.}
Firstly, we pre-train \textsc{Seagull} on \textsc{Seagull}-100w to enhance the quality feature extraction capabilities. Then, we fine-tune the pre-trained \textsc{Seagull} on \textsc{Seagull}-3k to adapt to authentic images. The settings for these two stages are as follows:

\textbf{Stage 1: Pre-training on the \textsc{Seagull}-100w.}
At this stage, we fix the image encoder and pre-train the linear projector, MFE and LLM on \textsc{Seagull}-100w. 
This pre-training enables \textsc{Seagull} to capture more robust ROI quality-related features, enhancing the efficiency of subsequent fine-tuning with smaller datasets.

\textbf{Stage 2: Fine-tuning on the \textsc{Seagull}-3k.} At this stage, we load the pre-trained weights from Stage 1 and fix the visual encoder while fine-tuning the remaining modules on \textsc{Seagull}-3k dataset. This stage allows \textsc{Seagull} to effectively adapt to real-world ROI-based IQA tasks.

\noindent \textbf{B) Implement Details.}
We use ConvNeXt\_large \cite{liu2022convnet} as the image encoder and the Vicuna \cite{chiang2023vicuna} as the LLM, with a maximum token length of 2,048. We apply the AdamW for training, with images resized to $512\times512$. 
Both stages are trained using LoRA \cite{hu2021lora} with lora\_rank of 128 and lora\_alpha of 256 on 8 NVIDIA 3090 GPUs, and the learning rate is adjusted using the Cosine Annealing. At the stage 1, pre-training is conducted with a batch size of 32 and a learning rate of $4\times10^{-5}$ for 1 epochs. AIRs and JIRs are generated randomly based on ground truth (GT). At the stage 2, fine-tuning uses the fixed AIRs and JIRs with batch size of 32 and learning rate of $1\times10^{-5}$ for 5 epochs. The images in \textsc{Seagull}-3k are split into 80\% for training and 20\% for testing to ensure no image overlapping.  Performance evaluations are based on the final models after 5 epochs on \textsc{Seagull}-3k. Following the Q-Align \cite{wu2023qalign}, we use closed-set probability calculations to translate discrete levels into continuous scores for all VLM-based methods on Quality Scores and Importance Scores.

\noindent \textbf{C) Evaluation Metrics.}
For ROI Quality Scores and Importance Scores, we use Spearman’s Rank Order Correlation Coefficient (SROCC) and Pearson’s Linear Correlation Coefficient (PLCC) for evaluation, both of which range from -1 to 1. For Distortion Severity Degrees and Distortion Types, we use Precision, Recall, and F1 score, all ranging from 0 to 1. The higher the values of all five metrics, the better the performance.

\subsection{Compared with State-of-the-art Models}
\noindent\textbf{A) ROI-based Assessment Comparison.}
\begin{table*}[h]
\huge       
\caption{ROI-based assessment comparison on four sub-tasks on the test set of \textsc{Seagull}-3k in terms of SROCC, PLCC, Sample-Average Precision, Sample-Average Recall and Sample-Average F1 Score. Best and second-best scores are marked in bold and underline, respectively. * denotes all-in-one models. $\dagger$ denotes pre-training on \textsc{Seagull}-100w.}
\label{tab:comparison_all}
\centering
\resizebox{\textwidth}{!}{
\begin{tabular}{c|c|cc|cc|ccc|ccc}
\toprule
 &  & \multicolumn{2}{c|}{\multirow{-1}{*}{\textbf{Quality  Score}}} & \multicolumn{2}{c|}{\multirow{-1}{*}{\textbf{Importance  Score}}} & \multicolumn{3}{c|}{\textbf{Distortion Severity Degree}} & \multicolumn{3}{c}{\textbf{Distortion Type Labels}} \\
\multirow{-2}{*}{\textbf{Models}} & \multirow{-2}{*}{\textbf{Inputs}} & \textbf{SROCC} & \textbf{PLCC} & \textbf{SROCC} & \textbf{PLCC} & \textbf{Precision (\%)} & \textbf{Recall (\%)} & \textbf{F1 Score (\%)} & \textbf{Precision (\%)} & \textbf{Recall (\%)} & \textbf{F1 Score(\%)} \\ \midrule
HyperIQA &  & 0.7120 & 0.7162 & 0.6645 & 0.6636 & — & — & — & — & — & — \\
DBCNN &  & 0.6836 & 0.6721 & 0.3832 & 0.3551 & — & — & — & — & — & — \\
QualiCLIP &  & 0.6166 & 0.6090 & 0.4902 & 0.4915 & — & — & — & — & — & — \\
PromptIQA* & \multirow{-4}{*}{Crop-based ROI} & {\ul 0.7377} & {\ul 0.7112} & 0.6028 & 0.5991 & — & — & — & — & — & — \\ \midrule
Yi-VL(6B)* &  & 0.5315 & 0.5427 & 0.6697 & 0.6926 & 21.07\% & 21.07\% & 21.07\% & 23.44\% & 23.44\% & 23.44\% \\
mPLUG-Owl2 (7B)* &  & 0.6281 & 0.6321 & 0.7176 & 0.7173 & {\ul 28.35\%} & 27.00\% & 26.69\% & 57.52\% & 56.37\% & 53.86\% \\
Qwen2-VL (7B)* &  & 0.6539 & 0.6533 & 0.7153 & 0.7161 & 27.41\% & 24.50\% & 25.02\% & 51.15\% & 45.03\% & 45.83\% \\
LLaVA-1.5 (7B)* &  & 0.5693 & 0.5774 & 0.7338 & 0.7377 & 25.10\% & 25.19\% & 24.14\% & {\ul 59.33\%} & 57.55\% & 54.95\% \\
mPLUG-Owl2 (Q-Align)* &  & 0.6562 & 0.6622 & 0.5339 & 0.5127 & 15.60\% & 12.20\% & 13.02\% & 52.44\% & 39.77\% & 42.19\% \\
mPLUG-Owl2 (Q-Instruct)* &  & 0.6644 & 0.6559 & 0.5172 & 0.5037 & 16.96\% & 25.25\% & 19.00\% & 40.80\% & {\ul 64.04\%} & 46.75\% \\
LLaVA-1.5 (Q-Instruct)* & \multirow{-7}{*}{\begin{tabular}[c]{@{}c@{}}BBox-based ROI\\ \& Full Image \& Text\end{tabular}} & 0.6606 & 0.6623 & 0.7667 & 0.7605 & 27.69\% & 26.52\% & 26.02\% & 57.87\% & 56.77\% & 53.96\% \\ \midrule
Osprey (7B)*$\dagger$ & Mask-based ROI & 0.7176 & 0.7173 & \textbf{0.8811} & \textbf{0.8756} & 27.17\% & {\ul 29.55\%} & {\ul 26.72\%} & 58.17\% & 62.52\% & {\ul 56.25\%} \\
\rowcolor[HTML]{E3F2D9} 
Seagull (7B)*$\dagger$ & \& Full Image \& Text & \textbf{0.7452} & \textbf{0.7465} & {\ul 0.8603} & {\ul 0.8468} & \textbf{29.50\%} & \textbf{32.51\%} & \textbf{29.03\%} & \textbf{59.90\%} & \textbf{66.87\%} & \textbf{59.08\%}
 \\\bottomrule
\end{tabular}
}
\end{table*}
To demonstrate the efficacy of \textsc{Seagull} in ROI quality analysis, we conduct experiments on four sub-tasks: ROI Quality Scores prediction, Importance Scores prediction, Distortion Severity Degrees prediction and Distortion Types identification. We compare the performance of \textsc{Seagull} with twelve SOTA models, namely four vision-based IQA models (HyperIQA \cite{su2020blindly}, DBCNN \cite{zhang2018blind}, QualiCLIP \cite{qualclip2024}, and PromptIQA \cite{chen2024promptiqa}), five VLMs (Yi-VL \cite{ai2024yi}, mPLUG-Owl2\cite{ye2023mplug}, Qwen2-VL \cite{Qwen2VL}, LLaVA-1.5 \cite{liu2024visual} and Osprey \cite{yuan2024osprey}) and three VLMs pre-trained on IQA related datasets (mPLUG-Owl2-Q-Align \cite{wu2023qalign}, mPLUG-Owl2-Q-Instruct \cite{wu2023qinstruct} and LLaVA-1.5-Q-Instruct  \cite{wu2023qinstruct}). Except for the HyperIQA, DBCNN and QualiCLIP, which train separate models for different sub-tasks, others train only one model for all sub-tasks (All-In-One). For the vision-based models, we adopt crop-based ROIs. For the VLMs, ROIs are indicated using BBoxes, except Osprey and \textsc{Seagull} using masks. The ROIs and text prompts are used to qurey the models for ROI quality analysis. We calculate Precision, Recall and F1 score between predictions and GTs for each ROI sample. Then, we average the results across all ROIs to obtain Sample-Average Precision, Sample-Average Recall and Sample-Average  F1 score.

As shown in Tab.\ref{tab:comparison_all}, \textsc{Seagull} outperforms most compared models on the four sub-tasks. For vision-based methods, which take the crop-based ROIs as the input, they show limited performance on Importance Score prediction. This is because the cropped ROIs discard global information, limiting the models' ability to understand the ROIs' impact on the overall image. Additionally, compared to VLMs that use BBoxes to indicate ROIs, \textsc{Seagull} surpasses them on all sub-tasks. The results suggest that BBox-based ROIs introduce irrelevant background information and fail to precisely indicate ROIs, leading to  inaccurate guidance. Despite these methods have been pre-trained on large-scale datasets, they still exhibit suboptimal performance. This is since that most datasets for VLM pre-training primarily focus on high-level tasks, resulting in a misalignment with the low-level task. Furthermore, VLMs pre-trained on IQA-related datasets, such as mPLUG-Owl2 (Q-Align), mPLUG-Owl2 (Q-Instruct) and LLaVA-1.5 (Q-Instruct), still show limited performance. This is because these datasets are designed for overall quality analysis, which differs from the fine-grained ROI-based IQA.
Compared to Osprey$\dagger$, which uses mask-based ROIs and is pre-trained on \textsc{Seagull}-100w, \textsc{Seagull}$\dagger$ demonstrates superior performance across most sub-tasks, excepting Importance Score prediction. This slight advantage for Osprey$\dagger$ (0.02 in SROCC and 0.03 in PLCC) can be attributed to its design for image content comprehension, enabling better ability of semantic understanding. Since Importance Score prediction highly depends on the semantics of the ROI and the global context, Osprey performs marginally better in this specific sub-task, 

It worth noting that all models' metrics in the Severity Degree identification task are generally low. In this task, a prediction is considered correct only when both the Severity Degree and the Distortion Type align with the GT. Thus, this is a multi-class task with 30 categories (5 severity degree for 6 distortion types), making it more challenging than other sub-tasks. Nevertheless, \textsc{Seagull} still surpasses the second-best model by approximately 4.05\% to 10.00\%.

\noindent\textbf{B) Distortion Type Identification Comparison.} In this experiments, we compare the F1 scores of \textsc{Seagull} with other VLMs across each distortion type. The results in Tab.\ref{tab:distrotion_types} demonstrate that \textsc{Seagull} outperforms other VLMs on nearly all distortion types and the clean condition (Without Distortion). On average, the F1 Score of \textsc{Seagull} surpasses that of the second-best model (Osprey) by 5.39\%. 
Some models which are not pre-trained on \textsc{Seagull}-100w struggle with Compression identification, primarily due to the smaller number of related samples in \textsc{Seagull}-3k, making this distortion more challenge to learn. 
Additionally, VLMs pre-trained on other IQA datasets, such as mPLUG-Owl2 (Q-Align) and mPLUG-Owl2 (Q-Instruct), exhibit limited performance in distortion identification. This is because these IQA datasets focus on overall quality assessment rather than fine-grained ROI-based distortion understanding. 
In contrast, Osprey$\dagger$ and \textsc{Seagull}$\dagger$, both pre-trained on \textsc{Seagull}-100w, demonstrate the capability to identify all distortion types effectively. These results further demonstrate the effectiveness of the \textsc{Seagull}-100w dataset and the efficiency of the \textsc{Seagull}.

\begin{table*}[]
\huge
\caption{Distortion types identification accuracy comparison on the test set of \textsc{Seagull}-3k in terms of F1 Score. Best and second-best scores are highlighted in bold and underline, respectively. $\dagger$ denotes pre-training on \textsc{Seagull}-100w.}
\label{tab:distrotion_types}
\resizebox{\textwidth}{!}{
\begin{tabular}{c|c|c|c|c|c|c|c|c|c}
\toprule
\textbf{Models} & \textbf{ROI Type} & \textbf{Blur} & \textbf{Colorfulness} & \textbf{Noise} & \textbf{Compression} & \textbf{Contrast} & \textbf{Exposure} & \textbf{Clean} & \textbf{Average} \\ \midrule
Qwen2-VL &  & 67.14\% & 14.93\% & 37.30\% & 0.00\% & 17.24\% & 38.16\% & \textbf{53.33\%} & 32.58\% \\
LLaVA-1.5 &  & 79.09\% & 33.63\% & 39.59\% & {\ul 23.53\%} & 23.91\% & {\ul 51.34\%} & 49.56\% & 42.95\% \\
mPLUG-Owl2 &  & 79.41\% & 22.70\% & 42.38\% & 9.52\% & 20.93\% & 47.88\% & 44.71\% & 38.22\% \\
mPLUG-Owl2 (Q-Align) &  & 69.38\% & 15.69\% & 24.03\% & 0.00\% & 18.39\% & 30.00\% & 37.78\% & 27.89\% \\
mPLUG-Owl2 (Q-Instruct) & \multirow{-5}{*}{\begin{tabular}[c]{@{}c@{}}BBox-based ROI\\ \& Full Image \& Text\end{tabular}} & 78.70\% & 33.02\% & 38.58\% & 5.26\% & 10.81\% & 47.45\% & 1.58\% & 30.77\% \\ \midrule
Osprey$\dagger$  & Mask-based ROI & {\ul 81.05\%} & {\ul 38.91\%} & {\ul 46.43\%} & 20.83\% & \textbf{ 28.57\%} & 50.10\% & 45.83\% & {\ul 44.53\%} \\
\rowcolor[HTML]{E3F2D9} 
Seagull$\dagger$  & \cellcolor[HTML]{E3F2D9}\& Full Image \& Text & \textbf{83.33\%} & \textbf{39.48\%} & \textbf{52.20\%} & \textbf{25.00\%} & {\ul 24.00\%} & \textbf{51.94\%} & {\ul 52.58\%} & \textbf{46.93\%}
\\ \bottomrule
\end{tabular}
}
\end{table*}

\subsection{More Discussion}
\noindent \textbf{A) Pre-training Impact Discussion.}
To evaluate the impact of pre-training with \textsc{Seagull}-100w on \textsc{Seagull}’s performance, we pre-train \textsc{Seagull} using various data scales (0\%, 25\%, 50\% and 100\%) and subsequently fine-tune it on \textsc{Seagull}-3k. The results in Tab.\ref{tab:data_num} indicate that \textsc{Seagull} pre-trained with only 25\% data significantly outperforms the model without pre-training (0\%) on all sub-tasks. This validates that pre-training on the \textsc{Seagull}-100w effectively enhances model's ability of quality feature extracting. Additionally, with pre-training scales increasing, \textsc{Seagull}'s performance on Importance Score, Distortion Severity Degree and Distortion Types consistently improves. However, as the pre-trianing scales up from 50\% to 100\%, the improvement of \textsc{Seagull} in Quality Score prediction slows down. This is due to the gap between the synthetic and authentic Dist. images, resulting in negligible performance gains even with more pre-training data.

\begin{table}[htbp]
\huge   
\caption{The impact of pre-training scales on \textsc{Seagull}-100w in terms of SROCC, PLCC, Sample-Average Recall and Sample-Average F1 Score. Best scores are highlighted in bold.}
\label{tab:data_num}
\resizebox{\columnwidth}{!}{
\begin{tabular}{c|cc|cc|cc|cc}
\toprule
\multirow{2}{*}{\textbf{Scale}} & \multicolumn{2}{c|}{\textbf{Quality Score}} & \multicolumn{2}{c|}{\textbf{Importance Score}} & \multicolumn{2}{c|}{\textbf{Distortion Degree}} & \multicolumn{2}{c}{\textbf{Distortion Type}} \\
 & \textbf{SROCC} & \textbf{PLCC} & \textbf{SROCC} & \textbf{PLCC} & \textbf{Recall} & \textbf{F1 Score} & \textbf{Recall} & \textbf{F1 Score} \\ \midrule
0\% & 0.6236 & 0.6238 & 0.7512 & 0.7628 & 28.09\% & 25.49\% & 55.94\% & 50.18\% \\
25\% & 0.6892 & 0.6866 & 0.7760 & 0.7776 & 28.10\% & 25.64\% & 61.64\% & 56.12\% \\
50\% & 0.7441 & 0.7389 & 0.7878 & 0.7926 & 30.34\% & 28.20\% & 64.79\% & 58.11\% \\
100\% & \textbf{0.7452} & \textbf{0.7465} & \textbf{0.8603} & \textbf{0.8468} & \textbf{32.51\%} & \textbf{29.03\%} & \textbf{66.87\%} & \textbf{59.08\%} \\
\bottomrule
\end{tabular}
}
\end{table}

\noindent  \textbf{B) Ablation Studies.}
To evaluate the efficiency of proposed components, we train four variants:
I) w/o Pre-train, which is directly trained  on \textsc{Seagull}-3k;
II) w/o JIR, which is trained on \textsc{Seagull}-3k that excludes JIRs as described in Sec.\ref{sec:instruction_response_design}; 
III) w/o Local View Tokens and IV) w/o Global View Tokens, which remove the local and global tokens from the MEF, respectively. Except the w/o Pre-train, other variants are pre-trained on \textsc{Seagull}-100w.

The results from Tab.\ref{tab:ablation} demonstrate that removing any components negatively impacts model performance. Notably, the variant without pre-training shows remarkable decline in performance, validating the significance of pre-training on the proposed \textsc{Seagull}-100w. Additionally, removing the JIR during training also shows a drop performance in the four sub-tasks, especially for the Quality Score and Importance Scores, suggesting that JIR enhances the \textsc{Seagull}'s understanding of these aspects. Furthermore, the removal of global tokens significantly degrades the performance, particularly in Importance Score prediction. This is because global tokens contain rich quality and semantic features. Moreover, the results of removing the local tokens indicate that the local detail provided by local view tokens enhance \textsc{Seagull}'s understanding of fine-grained details.

\begin{table}[htbp]
\huge                                                  
\caption{Ablation studies on critical components of the \textsc{Seagull} in terms of SROCC, PLCC, Sample-Average Recall and Sample-Average F1 Score. Best scores are highlighted in bold.}
\label{tab:ablation}
\resizebox{\columnwidth}{!}{
\begin{tabular}{c|cc|cc|cc|cc}
\toprule
\multirow{2}{*}{\textbf{Variants}} & \multicolumn{2}{c|}{\textbf{Quality Score}} & \multicolumn{2}{c|}{\textbf{Importance Score}} & \multicolumn{2}{c|}{\textbf{Severity Degree}} & \multicolumn{2}{c}{\textbf{Distortion Degree}} \\
 & \textbf{SROCC} & \textbf{PLCC} & \textbf{SROCC} & \textbf{PLCC} & \textbf{Recall} & \textbf{F1 Score} & \textbf{Recall} & \textbf{F1 Score} \\ \midrule
w/o Pre-train & 0.6236 & 0.6238 & 0.7512 & 0.7628 & 28.09\% & 25.49\% & 55.94\% & 50.18\% \\
w/o JIR & 0.6954 & 0.7022 & 0.8020 & 0.7874 & 31.37\% & 28.91\% & 63.59\% & 58.12\% \\
w/o Local & 0.7211 & 0.7331 & 0.8538 & 0.8409 & 31.49\% & 28.72\% & 65.44\% & 58.16\% \\
w/o Global & 0.5671 & 0.5761 & 0.2475 & 0.2503 & 27.04\% & 24.37\% & 62.14\% & 54.82\% \\ \midrule
Full & \textbf{0.7452} & \textbf{0.7465} & \textbf{0.8603} & \textbf{0.8468} & \textbf{32.51\%} & \textbf{29.03\%} & \textbf{66.87\%} & \textbf{59.08\%} \\
\bottomrule
\end{tabular}
}
\end{table}

\begin{figure}[htbp]
    \centering
    \includegraphics[width=\columnwidth]{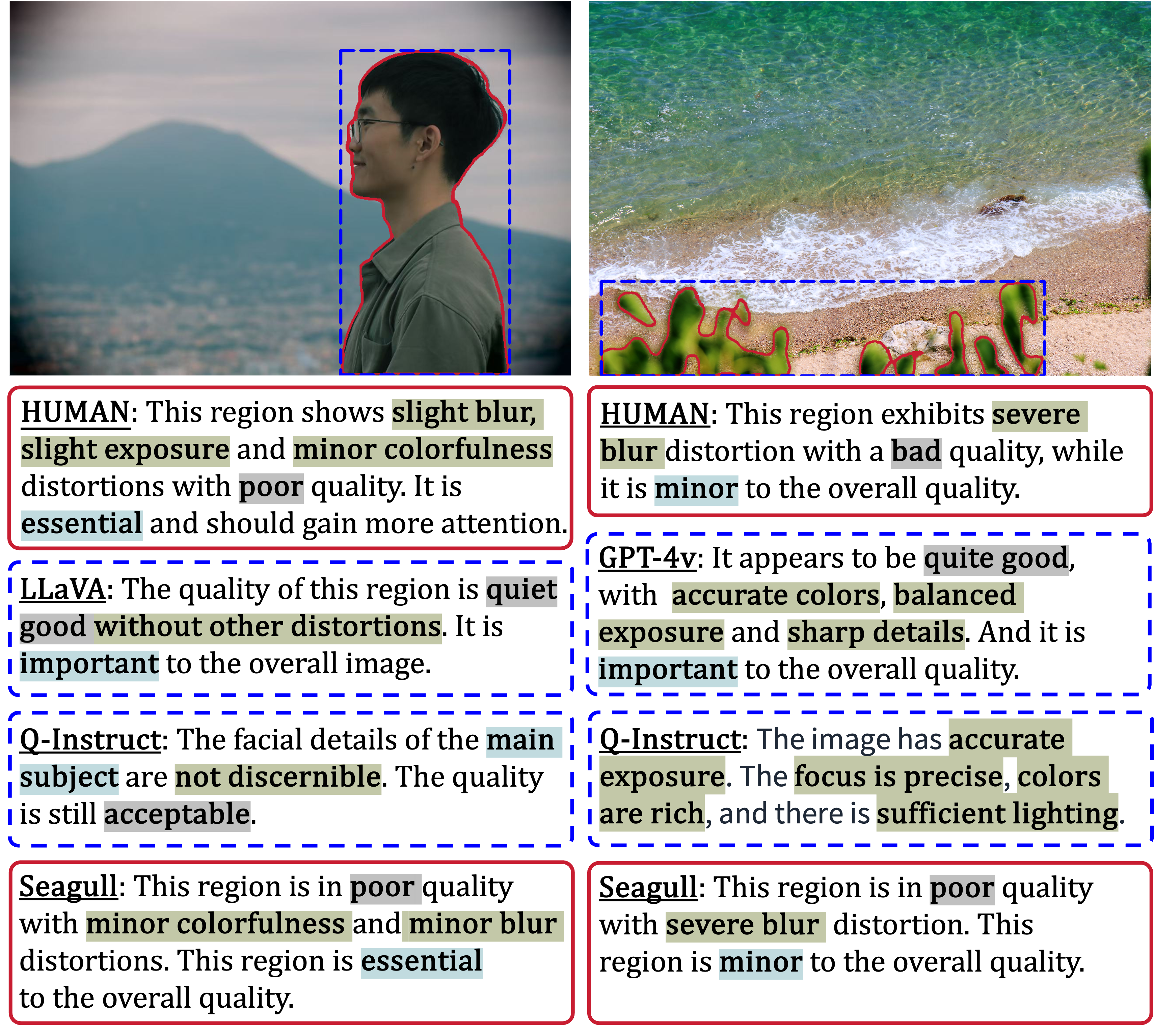}
    \caption{ROI quality analysis results from Human, VLMs and \textsc{Seagull}. Best viewed in color.}
    \label{fig:visualization}
\end{figure}

\noindent \textbf{C) Visualization.} We visualize analysis results from \textsc{Seagull}, human and three open-access VLMs: LLaVA-1.5, GPT-4v \cite{achiam2023gpt}, and Q-Instruct, using authentic images. As illustrated in Fig.\ref{fig:visualization}, all VLMs face challenges in analyzing the quality of ROIs. 
In contrast, \textsc{Seagull} demonstrates a strong consistency with human perception, which further validates it's reliability.

\section{Conclusion}
\label{sec:conclusion}

In this paper, we have achieved IQA for ROIs using VLMs and addressed two key challenges: the difficulty of VLMs in perceiving fine-grained quality and the lack of ROI-based IQA datasets. Firstly, we propose a novel network, \textsc{Seagull}, which incorporates a VLM, masks generated by SAM, and a carefully designed MFE. Then, we construct a large-scale synthetic Dist. dataset, \textsc{Seagull}-100w, for pre-training to make models effectively percept ROI quality. And we manually annotate an authentic Dist. dataset, \textsc{Seagull}-3k, for fine-tuning to improve models' generalization on authentic Dist. images. Experiments demonstrate that after pre-training on \textsc{Seagull}-100w and fine-tuning on \textsc{Seagull}-3k, \textsc{Seagull} exhibits outstanding ROI quality assessment capabilities. 

{
    \small
    \bibliographystyle{ieeenat_fullname}
    \bibliography{main}
}

\end{document}